\documentclass[sigconf]{acmart}

\usepackage{amsmath,amsfonts}
\usepackage{algorithmic}
\usepackage{graphicx}
\usepackage{textcomp}
\usepackage[normalem]{ulem}
\usepackage{hyperref}
\usepackage{booktabs}
\usepackage{multirow}
\usepackage{color, colortbl}
\usepackage{comment}

\AtBeginDocument{%
  \providecommand\BibTeX{{%
    \normalfont B\kern-0.5em{\scshape i\kern-0.25em b}\kern-0.8em\TeX}}}

\begin{document}

\title{Function Composition in Trustworthy Machine Learning: Implementation Choices, Insights, and Questions}

\author{Manish Nagireddy}
\affiliation{%
  \institution{IBM Research}
  \city{Yorktown Heights}
  \country{USA}}
\email{manish.nagireddy@ibm.com}

\author{Moninder Singh}
\affiliation{%
  \institution{IBM Research}
  \city{Yorktown Heights}
  \country{USA}}
\email{moninder@us.ibm.com}

\author{Samuel C. Hoffman}
\affiliation{%
  \institution{IBM Research}
  \city{Yorktown Heights}
  \country{USA}}
\email{shoffman@ibm.com}

\author{Evaline Ju}
\affiliation{%
  \institution{IBM Research}
  \city{Yorktown Heights}
  \country{USA}}
\email{evaline.ju@ibm.com}

\author{Karthi\-keyan Natesan Ramamurthy}
\affiliation{%
  \institution{IBM Research}
  \city{Yorktown Heights}
  \country{USA}}
\email{knatesa@us.ibm.com}

\author{Kush R. Varshney}
\affiliation{%
  \institution{IBM Research}
  \city{Yorktown Heights}
  \country{USA}}
\email{krvarshn@us.ibm.com}

\renewcommand{\shortauthors}{Nagireddy, et al.}

\begin{abstract}
  Ensuring trustworthiness in machine learning (ML) models is a multi-dimensional task. In addition to the traditional notion of predictive performance, other notions such as privacy, fairness, robustness to distribution shift, adversarial robustness, interpretability, explainability, and uncertainty quantification are important considerations to evaluate and improve (if deficient). However, these sub-disciplines or `pillars' of trustworthiness have largely developed independently, which has limited us from understanding their interactions in real-world ML pipelines. In this paper, focusing specifically on compositions of functions arising from the different pillars, we aim to reduce this gap, develop new insights for trustworthy ML, and answer questions such as the following. Does the composition of multiple fairness interventions result in a fairer model compared to a single intervention? How do bias mitigation algorithms for fairness affect local post-hoc explanations? Does a defense algorithm for untargeted adversarial attacks continue to be effective when composed with a privacy transformation? 
  Toward this end, we report initial empirical results and new insights from 9 different compositions of functions (or pipelines) on 7 real-world datasets along two trustworthy dimensions - fairness and explainability. We also report progress, and implementation choices, on an extensible composer tool to encourage the combination of functionalities from multiple pillars. To-date, the tool supports bias mitigation algorithms for fairness and post-hoc explainability methods.
  We hope this line of work encourages the thoughtful consideration of multiple pillars when attempting to formulate and resolve a trustworthiness problem.
\end{abstract}


\maketitle

\section{Introduction}

Machine learning (ML) has become ubiquitous, influencing outcomes in domains including, but not limited to, credit scoring, social media, public services, healthcare, education, and criminal justice. Consequently, a growing discipline of research, known as trustworthy or responsible ML, has drawn attention to the ways these systems lead to abject harm from safety and security issues, and exacerbate existing social inequities or even construct new ones \cite{Varshney2022}. Trustworthiness in machine learning takes many forms, including `pillars' such as privacy, fairness, robustness to distribution shift, robustness to adversarial attacks, interpretability, explainability, and uncertainty quantification. Metrics and methods developed in the field are used to detect and defend against harms by means of mitigation algorithms and transparency algorithms. In some form or other, all of these metrics and methods are mathematical functions applied to data.

Despite their co-occurrence in consequential domains and their inter-relatedness, much of the work on the different pillars of trustworthiness to-date has developed independently. In contrast, this paper starts to look at what happens when functions from different pillars (or even the same pillar) are composed together. 
The types of questions we are interested in answering are:
\begin{itemize}
\item Does the composition of multiple fairness interventions result in a fairer model compared to a single intervention?
\item How do bias mitigation algorithms for fairness affect local post-hoc explanations?
\item Should a calibration method be applied before or after a post-processing intervention?
\item Does a defense algorithm for untargeted adversarial attacks continue to be effective when composed with a privacy transformation?
\item Are directly interpretable models more or less able to be robustified against distribution shifts than uninterpretable models?

\end{itemize}
Toward our goal, we report initial
empirical results and new insights from 9 different compositions
of functions (or pipelines) on 7 real-world datasets along two
trustworthy dimensions - fairness and explainability. We also
report progress, and implementation choices, on an extensible
composer tool that allows users to string together functions from these dimensions to gain insights on intersectional trust-related questions. 

\begin{figure*}
\centerline{\includegraphics[width=0.5\textwidth]{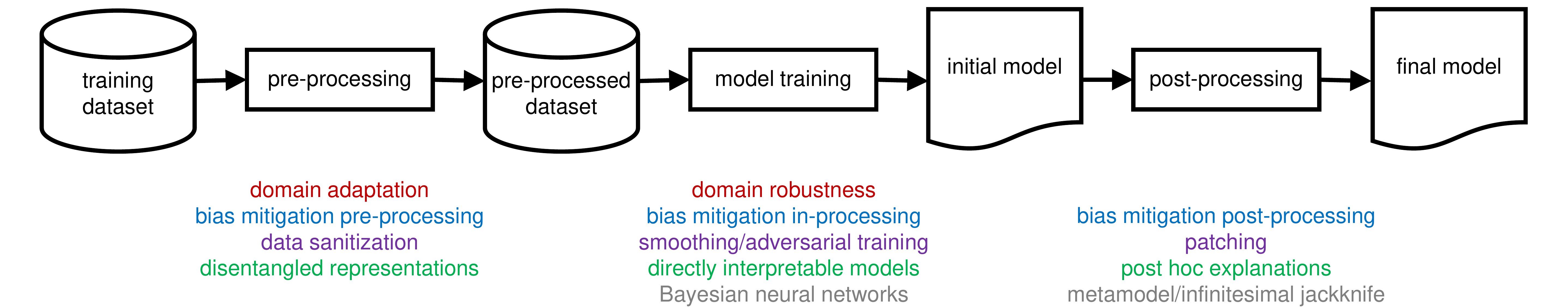}}
\caption{High-level ML pipeline with trustworthy functions at each step. Red functions are for robustness to distribution shift, blue functions are for fairness, purple functions are for adversarial robustness, green functions are for explainability or interpretability, and gray functions are for uncertainty quantification.}
\label{fig:bigpicture}
\end{figure*}

As illustrated in Fig.~\ref{fig:bigpicture}, functions arising in different pillars have common groupings of where they occur in an ML pipeline (pre-processing, model training, and post-processing), and thus lend themselves to being composed with each other within the same grouping or across groupings.
Thus far, we have focused on bias mitigation algorithms applicable in the three different parts of an ML pipeline and on post-hoc explanations, and are already able to obtain new insights. This has involved combining capabilities from Scikit-learn \cite{pedregosa2011scikit}, AI Fairness 360 \cite{aif}, Fairlearn \cite{bird2020fairlearn}, and AI Explainability 360 \cite{arya2020ai}, which has already necessitated several implementation choices. 

While we have initially restricted ourselves to these two dimensions due to the complexity of evaluating multiple dimensions with a large number of methods and metrics, we plan to extend our experimentation next to other dimensions, such as adversarial robustness and privacy,  and consequently keep adding to the composer with capabilities from various libraries such as Adversarial Robustness Toolbox \cite{nicolae2018adversarial}, Uncertainty Quantification 360 \cite{ghosh2021uncertainty}, and others.

The rest of this paper is organized as follows. Section \ref{sec:background} provides background knowledge and related work. Section \ref{sec:composer} presents the composer tool, lists its features, and discusses particular implementation choices that we made. Section \ref{sec:experiment} describes an initial set of experiments we conducted using the tool to answer some of the aforementioned  questions, specifically the effect of composing functions along fairness and explainability dimensions. Section \ref{sec:results} gives the results and insights gleaned from these experiments. Section \ref{sec:conclusion} concludes and discusses future work.
\section{Background and Related Work}
\label{sec:background}

Trustworthy ML is a vast topic with entire books written about individual pillars \cite{LiLLCL2022,barocas-hardt-narayanan,ChenH2023,molnar2022}. Rather than recounting this entire literature, we provide a brief review of the two pillars with which we have started our effort: fairness and explainability. We also describe prior work on large-scale empirical studies in trustworthy ML, but note that they are usually conducted on single pillars or do not involve the composition of functions as in this work.

\subsection{Fairness}

First, we briefly define specialized terminology from the field of fairness in machine learning. \textit{Bias} is a systematic error. In the context of fairness, we are concerned with unwanted bias that places privileged groups at a systematic advantage and unprivileged groups at a systematic disadvantage. A \textit{protected attribute} is a feature from the data that partitions a population into groups that should have parity in terms of benefit received, and delineates privileged and unprivileged groups. Examples include race, gender, and religion. Protected attributes are not universal, but are application specific, and are also referred to as sensitive attributes. A \textit{favorable label} is a desirable outcome, like getting hired or not being arrested.

\subsubsection{Metrics}

A group \textit{fairness metric} is a quantification of unwanted bias in the machine learning pipeline. A key metric is \textit{statistical parity difference}, given by $$Pr(\hat Y = + \mid D = unpriv) - Pr(\hat Y = + \mid D = priv)$$ where $\hat{Y} = +$ indicates the predicted label being favorable, $D$ is the sensitive attribute, and $unpriv$ and $priv$ represent unprivileged and privileged classes, respectively. A value close to $0$ is desired; larger absolute values (either positive or negative) imply unfairness. Similarly, we have the \textit{disparate impact ratio} (DI), given by $$\frac{Pr(\hat Y = + \mid D = unpriv)}{Pr(\hat Y = + \mid D = priv)}.$$ A value of $1$ is desired.

Other group fairness metrics include \textit{equal opportunity difference}, which computes the difference in true positive rates between the unprivileged and privileged groups, and the \textit{average odds difference}, which computes the average of the difference in false positive rates and true positive rates between the unprivileged and privileged groups. In both cases, a value close to $0$ indicates fairness. The choice among fairness metrics is nuanced and discussed in \cite{Varshney2022}.

\subsubsection{Algorithms}

A bias mitigation algorithm is a procedure for reducing unwanted bias in an ML system. Broadly, there are three classes of fairness algorithms:

\begin{itemize}
    \item \textit{Pre-processing} algorithms modify the training data to reduce the underlying biases. These algorithms are best suited when training data is accessible and modifiable. 
    \item \textit{In-processing} algorithms reduce biases during the training process. These methods are best suited when there is direct access to the model training.
    \item \textit{Post-processing} algorithms modify model output predictions. These are best suited when data and model are both inaccessible -- they only requires access to black-box predictions. 
\end{itemize}

\subsubsection{Toolkits}

Recent years have seen the development of many open-source ML fairness toolkits intended to assist practitioners in assessing and reducing unfairness in ML systems. 
Two toolkits that provide fairness metrics, as well as extensive support for bias mitigation, are AI Fairness 360 (AIF360) \cite{aif} and Fairlearn \cite{bird2020fairlearn}. 

\subsection{Interpretability / Explainability}

Although understanding the data (e.g. extracting representative prototypes) can also be important, in this work, we focus on \textit{model explainability}, i.e., making the model comprehensible to the consumer. We also focus our initial work on \textit{post-hoc explanations} which utilize surrogate models for \textit{black box models} (uninterpretable models) as opposed to \textit{directly interpretable models}. Finally, explanations can be \textit{local} (explain one instance at a time) or \textit{global} (explain the entire model). 

\subsubsection{Metrics}

The ultimate evaluation for explainability should be user evaluations by people from the persona who the explanations are intended for \cite{doshi2017towards}. Nevertheless, quantitative metrics are often used to evaluate explainability. One of the main such metrics is \emph{faithfulness}, which evaluates the correlation between the feature importance assigned by an interpretability algorithm and the effect of each of these features on the predictive performance of the model \cite{alvarez2018towards}. Intuitively, the higher the importance, the higher should be the effect, and vice versa. The metric evaluates this by incrementally removing each of the features deemed important by the interpretability metric, evaluating the effect on the performance, and then calculating the correlation between the weights (or importance) of the attributes and the corresponding model performance. 

\subsubsection{Algorithms}
As ML models have become increasingly complex and many model providers keep model details proprietary, users have to treat these models as black boxes. Users may want to understand what led models to make a certain decision before they take further action or use the models in larger systems. 
Post-hoc explanation methods find relationships between features and outcomes to provide explanations on black box models. Prominent techniques such as SHAP \cite{NIPS2017_8a20a862}, LIME \cite{ribeiro2016should}, and CEM \cite{dhurandhar2019model} provide local, model-agnostic explanations.

\subsubsection{Toolkits}

Several open-source toolkits for explainability have been released over the past few years. Among them are AI Explainability 360 (AIX360) \cite{arya2020ai}, InterpretML \cite{nori2019interpretml}, and OpenXAI \cite{agarwal2022openxai}.

\subsection{Prior Empirical Studies}

A number of empirical studies have been carried out that discuss the relative merits of different methods within one pillar as well as attempt to understand the effect of interaction of methods across pillars. To set the stage for our work, we only focus on quantitative studies that only minimally involve the human component, if at all. Note that in this discussion, we explicitly mention \textit{data pre-processing} when we refer to standard feature engineering and data preparation techniques, whereas just \textit{pre-processing} implies pre-processing techniques used for bias mitigation. The main methodological difference between our work and the following is that we provide a flexible way to compose various trust functions. 

\subsubsection{Fairness}
Engineering practices used in the ML pipeline for data processing such as test data isolation and hyperparameter tuning can significantly impact the effect of fairness-enhancing interventions. In \cite{schelter2019fairprep}, the authors demonstrate this and discuss how the variability in such interventions can be reduced by using good practices for data-handling in the pipeline. In \cite{feffer2022empirical}, recognizing that the effect of bias mitigation methods can be unstable across data splits, the authors investigate and show that they can be used along with ensembling approaches to improve stability. They also release an open-source library that can be used to perform such studies. A comprehensive study of bias mitigation methods with multiple datasets along with a quantification of fairness-accuracy trade-off is performed in \cite{chen2022comprehensive}. Their results are a mixed bag, both in terms of fairness enhancement as well as the trade-off; their main conclusion is that no one method works well for all scenarios they considered. In a related vein, \cite{biswas2020machine} use top-rated models from the well-known Kaggle platform for multiple tasks and illustrate the advantage of pre-processing bias mitigation methods in terms of fairness-accuracy trade-off, unreliability of in-processing in preserving accuracy, and competitiveness of the post-processing methods again with respect to the trade-off. The authors in \cite{aif_arxiv} do a benchmarking study on various bias mitigation techniques with multiple datasets using the different fairness metrics and effectively demonstrate good improvement in fairness with a minimal reduction in accuracy for pre-processing, and the advantage of post-processing when only black-box model access is available. The authors also illustrate the correlation between multiple fairness metrics. Finally, \cite{friedler2019comparative} also present a benchmarking study where they show a similar correlation between metrics, along with variability in performance of bias mitigators with respect to data splits and data pre-processing. Overall, the studies hint at the following: (a) good engineering practices need to be followed when constructing trust pipelines since they are typically non-robust, (b) there is no single method that works well for all use cases, (c) pre-processing methods may have some benefit over others, (d) post-processing can be a convenient solution when only black-box access is available, (e) function composition of trust methods is still under-explored.

\subsubsection{Explainability}
Several empirical evaluations have been performed for explainability techniques using various metrics and use cases, although the studies are not as comprehensive as for fairness. LIME and SHAP are compared for two medical imaging use cases in \cite{hailemariam2020empirical}, showing that SHAP performs slightly better than LIME with respect to the metrics they considered. For the defect prediction use case, \cite{jiarpakdee2020empirical} shows that some model-agnostic (black-box) explanations can be efficient and reliable, except for LIME which can be less reliable. The extent to which different post-hoc explanations (LIME, SHAP, gradient-based methods) differ, using a notion of disagreement formalized based on inputs from stakeholders is discussed in \cite{krishna2022disagreement}. The authors show that the methods often disagree in unpredictable ways. Although explainability metrics are not as well-established as those for fairness, these works seem to point out that even when used alone, there is not a single explanation technique that can work well for all use cases and any approach that is chosen needs to be carefully validated for a given use case.

\subsubsection{Multiple Pillars}
In addition to just fairness or explainability, there are also works that study other trust pillars and a combination of multiple pillars. Ref.\ \cite{singh2021empirical} studies the performance of multiple classification models on eight different datasets using accuracy/balanced accuracy, and measures for fairness, explainability, adversarial robustness, and distributional robustness. The authors show that no single model type performs well on all these measures and demonstrate the various kinds of trade-offs involved. Out-of-domain faithfulness of post-hoc explanation methods evaluated using commonly used metrics is studied in \cite{chrysostomou2022empirical} and the authors claim that these metrics can be misleading in out-of-domain settings. The fairness and accuracy of models trained using invariant risk minimization (IRM) \cite{arjovsky2019invariant}, a recent technique for training distributionally robust models, is shown to be greater than those learned using empirical risk minimization, the most common approach for training ML models in \cite{adragna2020fairness}. In \cite{benz2021robustness}, the discrepancy in accuracy between classes (which is referred to as a form of unfairness) is shown to be aggravated with adversarial training hence suggesting that fairness and adversarial robustness may be at odds with each other. The relationship between accuracy and adversarial robustness is also discussed in \cite{su2018robustness} and several factors (including model architecture and network depth) that modulate this relationship for image classification are outlined. Finally, \cite{lyu2020differentially} shows that incorporating privacy guarantees via their training process also helps with model fairness over various demographic variables.  

Although these empirical studies that we have detailed above, and others such as \cite{agarwal2020trade,augustin2020adversarial,baniecki2021dalex,begley2020explainability,franco2021toward,grabowicz2022marrying,noack2021empirical}, compute metrics from more than one pillar of trustworthiness, they do not consider more than one algorithm or data transformation at a time. Observing metrics at the end of a pipeline is much simpler than implementing and understanding the composition of methods.

\section{Composer Tool}
\label{sec:composer}

\begin{figure*}[!h]
\centerline{\includegraphics[width=0.55\textwidth]{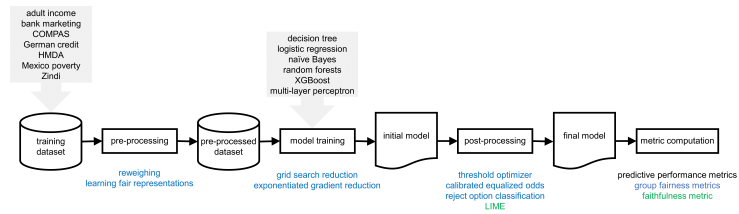}}
\caption{ML pipeline with currently implemented functionality at each step.}
\label{pipeline}
\end{figure*}

The proposed composer tool presents a framework to identify and interpret the effects of combining functionalities from different pillars of trustworthiness in (binary) classification settings. The tool can be visualized in terms of a simplified ML pipeline (Fig.~\ref{pipeline}). The composer incorporates modularity, where each module can correspond to a functionality, allows users to experiment with different configurations for the modules,
and provides choices at every stage of the ML pipeline. Specifically, any combination of methods can be selected, and the tool outputs both metrics as well as local explanations. 

\subsection{Implementation Choices}

We now elaborate on some implementation and design choices in order to further explain the details of our composer's capabilities.\footnote{We plan on open-sourcing our code through the AIF360 and AIX360 toolkits. Several of the authors are core developers and maintainers of these packages.} Due to the composer's extensibility, these capabilities can be easily expanded along multiple dimensions such as including more machine learning classification models, or including other pre-/in-/post-processing interventions.

\subsubsection{Configuration File}

Users are able to specify the exact desired composition via an intuitive YAML configuration file. First, at the data level, users specify the dataset and protected attribute of choice. Then, they can also provide the training, validation, and test set proportions.

Next, at the model level, users can choose up to five different machine learning models, including \texttt{Decision Tree}, \texttt{Logistic Regression}, \texttt{Naive Bayes}, \texttt{Random Forest}, \texttt{XGBoost} \cite{xgb}, and \texttt{Multi Layer Perceptron}. Users can also specify desired hyper-parameters for any of the above models.

Finally, at the pipeline level, users can list up to one intervention from any of the three classes of fairness algorithms (pre-, in-, and post-processing). Again, they may add algorithm-specific hyper-parameters if they so choose. Users can also elect to not use any fairness interventions (to serve as a baseline).

Specifying the configuration in this way allows users to experiment with different combinations of pipelines and automatically generate results to enable well-informed comparisons. For example, users can easily perform an ablation study of a complex pipeline by duplicating the configuration file and deleting different lines.

\subsubsection{Bias Mitigation Algorithms}

We enumerate the set of fairness algorithms that users can choose from.

\begin{itemize}
    \item \texttt{Reweighing (Rew)} \cite{rew} is a \textit{pre-processing} technique that weights the examples in each (group, label) combination differently to ensure equal group representation in the dataset.
    \item \texttt{Learning Fair Representations (LFR)} \cite{lfr} is a \textit{pre-processing} technique that finds a latent representation which encodes the data well but obfuscates information about protected attributes.
    \item \texttt{Grid Search Reduction (GridSearch)}~\cite{reductions} is an \textit{in-} \textit{processing} technique that reduces fair classification to a sequence of cost-sensitive classification problems, returning the deterministic classifier with the lowest empirical error subject to fair classification constraints.
    \item \texttt{Exponentiated Gradient Reduction (ExpGrad)} \cite{reductions} is an \textit{in-processing} technique that reduces fair classification to a sequence of cost-sensitive classification problems, returning a randomized classifier with the lowest empirical error subject to fair classification constraints.
    \item \texttt{Calibrated Equalized Odds (CEOdds)} \cite{ceodds} is a \textit{post-processing} technique that optimizes over calibrated classifier score outputs to find probabilities with which to change output labels for a relaxed equalized odds objective.
    \item \texttt{Reject Option Classification (ROC)} \cite{roc} is a \textit{post-processing} technique that gives favorable outcomes to unprivileged groups and unfavorable outcomes to privileged groups in a confidence band around the decision boundary with the highest uncertainty.
    \item \texttt{Threshold Optimizer (ThreshOptim)} \cite{hardt2016equality} is a \textit{post-} \textit{processing} technique that applies group-specific thresholds to the provided estimator. The thresholds are chosen to optimize the provided performance objective subject to the provided fairness constraints.
\end{itemize}

Thus, 2 pre-processing, 2 in-processing, and 3 post-processing fairness methods are available to be combined in the composer tool. 

Given that pipelines are created by composing multiple algorithms, each with their own customizable parameters, one needs to be cognizant of potential incompatibilities. Numerous works have pointed out existing issues with fairness definitions and metrics, both mathematically and philosophically \cite{heidari2019moral, friedler2016possibility}. Thus, when composing methods, it is imperative that choices of metrics and optimizers are consistent and compatible. For example, if \texttt{GridSearch} is composed with \texttt{ROC}, the constraint for \texttt{GridSearch} should be set to \texttt{DemographicParity} (as opposed to \texttt{EqualizedOdds}) if disparate impact is used as the scoring metric for \texttt{ROC}.

\subsubsection{Post-Hoc Explainability Algorithm}

We include an implementation of local interpretable model-agnostic explanations (LIME) \cite{ribeiro2016should}, a post-hoc explainability technique designed to explain individual predictions of any classifier in an interpretable manner, by learning a directly interpretable model (e.g., a linear model) locally around each prediction. Specifically, LIME works by estimating feature attributions on individual data instances, which capture the contribution of each feature on the black box prediction \cite{fool}.

In summary, the composer currently contains 7 fairness algorithms, one post-hoc explainability method, and an architecture that can be easily expanded with additional algorithms across pillars of trustworthiness.

\subsection{Example}
\label{sec:example}

To illustrate the tool's behavior, let us walk through an example. If a user desired to string together \texttt{Reweighing} (a pre-processing intervention) with \texttt{Grid Search Reduction} (an in-processing intervention) over \texttt{Logistic Regression} models (a base classifier) and \texttt{Reject Option Classification} (a post-processing intervention), the composer would work as follows.

It will first partition the data (which the user selects from a variety of datasets --- more specifics are provided in Section~\ref{sec:datasets}) into training, validation, and testing, using specific split proportions. Then, the composer will employ \texttt{Reweighing} on the training data and fit a machine learning model of the user's choice --- \texttt{Logistic Regression}, in this case. After this, the tool tunes the optimal classification threshold with respect to balanced accuracy on the validation data and uses this computed threshold to generate predictions on the test set. At this point, we have finished the pre-processing step and so the composer collects metrics (regarding performance, fairness, and explainability) as well as produces local explanations via LIME. We elaborate more on the specific metrics collected and explanations generated in Section~\ref{sec:experiment} regarding our experiments.

Next, the composer will proceed to employ the \texttt{Grid Search Reduction} method. Given that this in-processing intervention requires access to a base model, the supplied model will again be \texttt{Logistic Regression} and the overall meta-model will be trained on the \textit{reweighed data} from the pre-processing step. This represents the \textit{composition} of multiple fairness interventions. Again, we fit the model, compute the optimal threshold on the validation data, generate predictions on the test data, and report metrics as well as local explanations.

Finally, the composer will utilize \texttt{Reject Option Classification}. Functionally, this works similarly to the in-processing step in that we require a base model on which to post-process the predictions. In this case, the trained model from the \texttt{Grid Search Reduction} in-processing step will be used. However, for post-processing, the base model is trained separately on its own data split from the post-processing model itself (in contrast to \texttt{Grid Search Reduction} which can be thought of as a training procedure for the base model).

Hence, this represents the second and final composing of fairness interventions in our three-stage pipeline. In mathematical terms, this can be thought of as $\mathtt{Reject Option Classification} \circ (\mathtt{Grid Search} \circ \mathtt{Logistic Regression}) \circ \mathtt{Reweighing}$. The post-processing method will use an in-processing model which itself was trained on a pre-processed dataset.
Again, we proceed with the same steps of fitting, thresholding, predicting, and generating metrics and explanations.

After this experiment has completed, the user is able to compare three different sets of metrics and explanations: after pre-processing, after pre- and in-processing, and after pre-, in-, and post-processing. Additionally, if the user wishes to compare these interventions in isolation (e.g. what happens after only doing a post-processing algorithm), or if the user desires a baseline (i.e. no fairness algorithms at all), this is also possible by configuring our composer.

\section{Experiments}
\label{sec:experiment}

\subsection{Datasets}
\label{sec:datasets}

We used seven datasets described below and summarized in Table~\ref{tab1} in the supplemental material.

\emph{Adult Income} The Adult dataset \cite{d0} contains data on 48,442 individuals, with both demographic and financial information taken from the 1994 US Census Bureau database. The label, for prediction, is given by whether an individual will have an annual income of over \$50,000. Both `race' (`White' is privileged and `non-White' is unprivileged) and `sex' (`male' is privileged and `female' is unprivileged) are used as sensitive attributes.

\emph{Bank Marketing} The Bank dataset \cite{d1} contains data on 43,354 individuals related with direct marketing campaigns (e.g. phone calls) of a Portuguese banking institution. The label delineates whether a client will subscribe to a term deposit. We use `age' (`under 25' is privileged) as a sensitive attribute.

\emph{COMPAS Recidivism} The COMPAS dataset \cite{d2} contains data on 5,278 individuals, with both demographic and criminal history information of defendants from Broward County, Florida, from 2013--2014. The task is to predict whether a defendant will re-offend (i.e. recidivate) within two years. Both `race' (`Caucasian' is privileged and `African-American' is unprivileged) and `sex' (`female' is privileged and `male' is unprivileged) are used as sensitive attributes. Note that the favorable label in this dataset is 0 (i.e. defendant predicted to \textit{not} re-offend).

\emph{German Credit} The German Credit dataset \cite{d3} contains creditworthiness data on 1,000 individuals. The label represents whether an individual has good credit risk. Both `age' (`25 and older' is privileged) and `sex' (`male' is privileged and `female' is unprivileged) are used as sensitive attributes. 

\emph{Home Mortgage Disclosure Act (HMDA)} The HMDA dataset \cite{d5} contains data on 1,119,631 applications for single-family, principal-residence purchases, given from the 2018 mortgage application data collected in the U.S. under the Home Mortgage Disclosure Act. The label is predicting whether a mortgage is approved. We use `race-ethnicity` (`non-Hispanic Whites' are privileged and `non-Hispanic Blacks` are unprivileged) as the sensitive attribute.

\emph{Mexico Poverty} The Mexico Poverty dataset \cite{d4} contains data on 70,305 Mexican households, given from the 2016 Mexican household survey. The task is to predict whether a family is impoverished. We use `age' (given by the `young` feature) as a sensitive attribute.

\emph{Financial Inclusion in Africa (Zindi)} The Zindi dataset \cite{d6} contains data on 23,524 individuals, with both demographic and financial services information. The label represents whether an individual is likely to own a bank account. We use `gender' (given by the `gender-of-respondent' feature) as the sensitive attribute.

\subsection{Explanations}

We use LIME to generate post-hoc explanations as a final composition at the end of each pipeline to see the effect of composing fairness interventions on explanations. The explanations consist of feature importances for 10 individual samples randomly selected from the dataset but held constant across pipelines. These explanations are used in computing the faithfulness metric as well as detailed analysis of individual examples.

\subsection{Pipelines}

We ran 9 pipelines, or combinations, of fairness-enhancing algorithms, summarized in Table \ref{tab2}. As described in Section~\ref{sec:example}, for each pipeline, the tool will compute metrics after every step so shorter pipelines starting from the first step in each row are also implied. Additionally, we include a ``no fairness'' baseline where no fairness interventions are applied.

\definecolor{Gray}{gray}{0.95}

\begin{table}[htbp]
\caption{Summary of pipelines used in experiments}
\begin{center}
\scalebox{0.6}{
\begin{tabular}{ccccccccc}
\toprule
\multirow{2}{*}{Pipeline} & \multicolumn{2}{c}{Pre-processing} & \multicolumn{2}{c}{In-processing} & \multicolumn{3}{c}{Post-processing} & Explanations \\
\cmidrule(lr){2-3} \cmidrule(lr){4-5} \cmidrule(lr){6-8} \cmidrule(lr){9-9}
& \texttt{Rew} & \texttt{LFR} & \texttt{GridSearch} & \texttt{ExpGrad} & \texttt{CEOdds} & \texttt{ROC} & \texttt{ThreshOptim} & \texttt{LIME} \\
\midrule
1 & \checkmark & & \checkmark & & & \checkmark & & \checkmark \\
\rowcolor{Gray}
2 & \checkmark & & & \checkmark & & \checkmark & & \checkmark \\
3 & \checkmark & & & & \checkmark & & & \checkmark \\
\rowcolor{Gray}
4 & \checkmark & & & & & \checkmark & & \checkmark \\
5 & \checkmark & & & & & & \checkmark & \checkmark \\
\rowcolor{Gray}
6 & & \checkmark & & & \checkmark & & & \checkmark \\
7 & & \checkmark & & & & & \checkmark & \checkmark \\
\rowcolor{Gray}
8 & & & \checkmark & & & \checkmark & & \checkmark \\
9 & & & & \checkmark & & \checkmark & & \checkmark \\
\bottomrule
\end{tabular}
}
\label{tab2}
\end{center}
\end{table}

Each pipeline was run on the 10 dataset and protected attribute combinations mentioned above. Additionally, for each of these 10 configurations, we utilized 6 model classes: Decision Tree (DT), Logistic Regression (LR), Naive Bayes (NB), Random Forest (RF), and XGBoost (XGB), and a multi-layer perceptron (MLP).

\section{Results}
\label{sec:results}

\begin{figure}
\centerline{\includegraphics[scale=0.35]{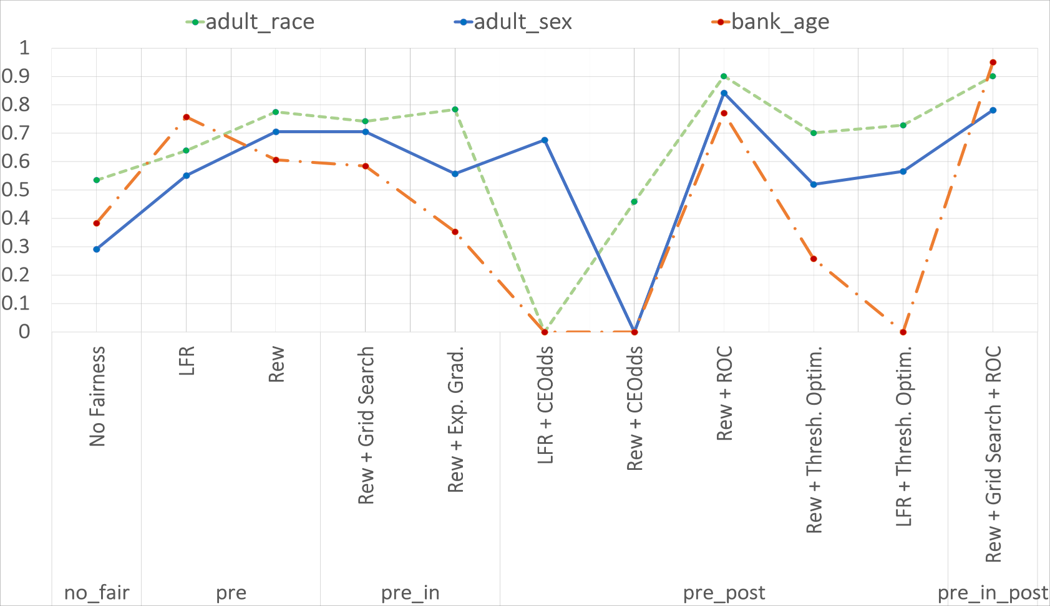}}
\caption{Disparate impact (DI) values for an XGBoost model after different types of fairness interventions involving a pre-processing fairness mitigation stage where DI value of base model before bias mitigation $<$ 0.6 and pre-processing bias mitigation resulted in a model with DI $<$ 0.8.}
\label{pre_low_DI_a}
\end{figure}

\begin{figure}
\centerline{\includegraphics[scale=0.35]{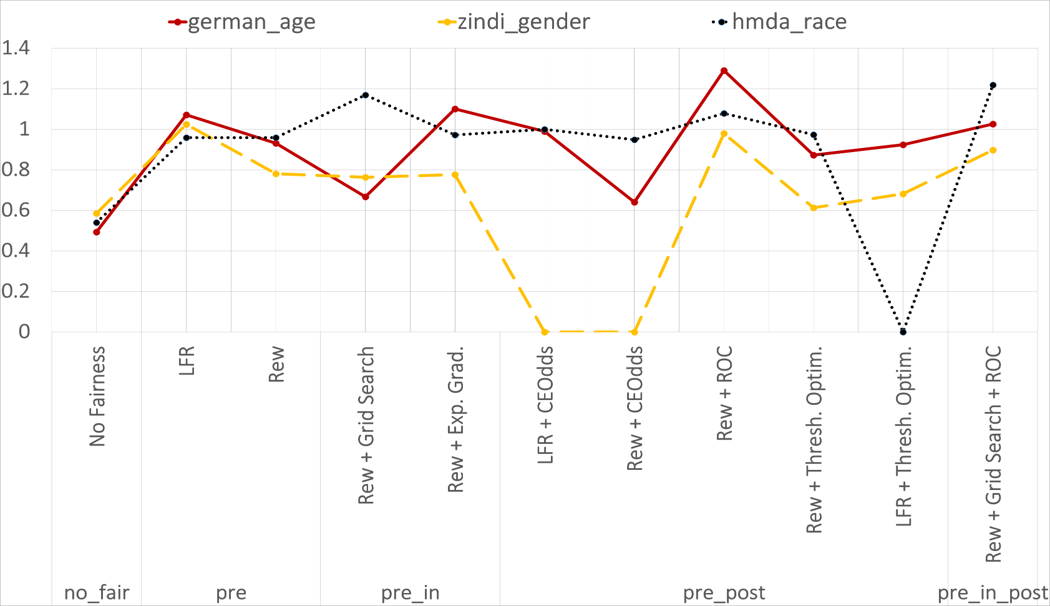}}
\caption{Disparate impact (DI) values for an XGBoost model after different types of fairness interventions involving a pre-processing fairness mitigation stage where DI value of base model before bias mitigation $<$ 0.6 and pre-processing bias mitigation resulted in a model with DI $>$ 0.8.}
\label{pre_low_DI_b}
\end{figure}

\begin{figure}
\centerline{\includegraphics[scale=0.35]{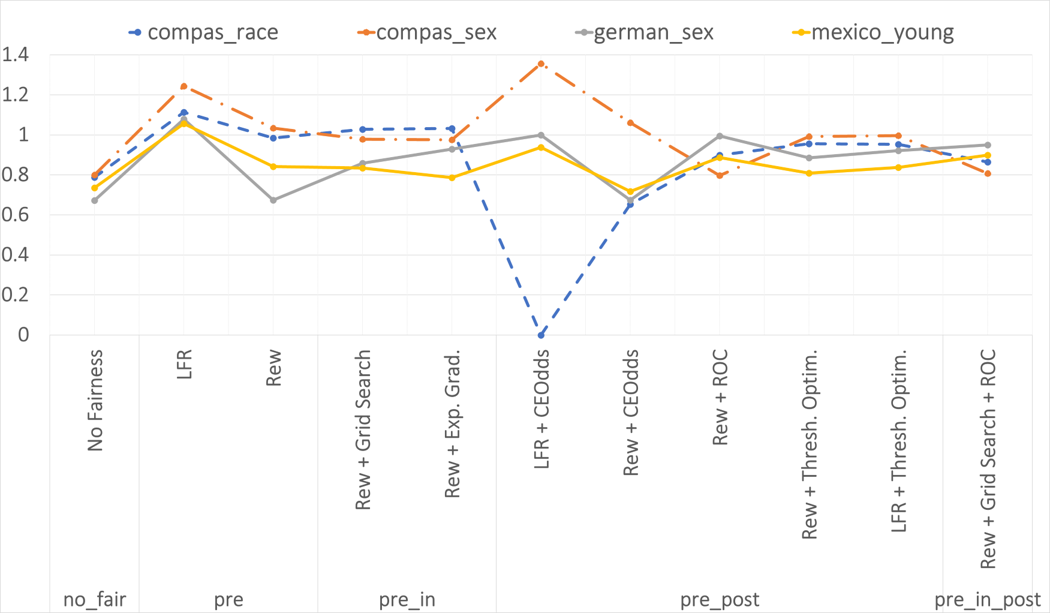}}
\caption{Disparate impact (DI) values for an XGBoost model after different types of fairness interventions involving a pre-processing bias mitigation stage where DI value of base model before bias mitigation $>$ 0.6.}
\label{pre_high_DI}
\end{figure}

In this section, we discuss the results of experiments involving the composition of multiple fairness interventions while learning a model.
For the sake of simplicity, we categorize the pipelines into two groups: pipelines which involve a pre-processing mitigation stage, and pipelines that only involve in-processing or post-processing interventions, i.e., no pre-processing.

\subsection{Pre-Processing Pipelines}
Fig. \ref{pre_low_DI_a}--\ref{pre_high_DI} show how the fairness of an XGBoost model learnt from various datasets is affected by compositions of bias mitigation interventions in pipelines that include a pre-processing step. 

Fig. \ref{pre_low_DI_a} shows datasets for which the base classifier was quite unfair (disparate impact (DI) $<$ 0.6). Moreover, despite the fact that pre-processing interventions (both \texttt{LFR} and \texttt{Rew}) improved the fairness of the models, the resultant models still had a DI value $<$ 0.8. Adding an additional in-processing intervention (\texttt{GridSearch} as well as \texttt{ExpGrad}) after \texttt{Reweighing} did not show any improvement, nor did the additional of \texttt{ThreshOptim} as a post-processing intervention. In these cases, the fairness actually deteriorated slightly in the case of the Bank dataset. However, the addition of \texttt{ROC} as a post-processing step increased the fairness of the resultant model each time, both with or without an intervening in-processing mitigation step. Addition of \texttt{CEOdds} as a post-processing step also did not show improvement; in fact, it often failed, primarily due to the relatively strict requirements requiring model calibration and error rate parity that could be difficult to achieve in practice.

Similar inferences can be drawn from Fig. \ref{pre_low_DI_b} that shows results for datasets that similarly had models that were quite unfair (DI $<$ 0.6) but the application of a pre-processing step typically resulted in models that were almost fair (DI close to 1). However, in these cases, a subsequent intervention often actually resulted in an over-correction, i.e., models that were made almost fair after pre-processing were made substantially unfair in the opposite direction (against the privileged group), both in the case of in-processing (\texttt{GridSearch} and \texttt{ExpGrad}) as well as post-processing interventions (\texttt{ROC}). Once again, \texttt{CEOdds} as well as the combination of \texttt{LFR} and \texttt{ThreshOptim} produced unexpected results in some cases.

Fig. \ref{pre_high_DI} shows datasets for which the base classifier was relatively less unfair (DI $>$ 0.6). In such cases, pre-processing typically worked  well in making the model fairer. However, \texttt{LFR} had a much stronger corrective effect compared to \texttt{Reweighing} that resulted in an over-correction leading to models that were biased in the other direction (i.e. DI $>$ 1). Adding on an additional intervention, in-processing or post-processing, typically has little effect in such cases.

\begin{figure}
\centerline{\includegraphics[scale=0.35]{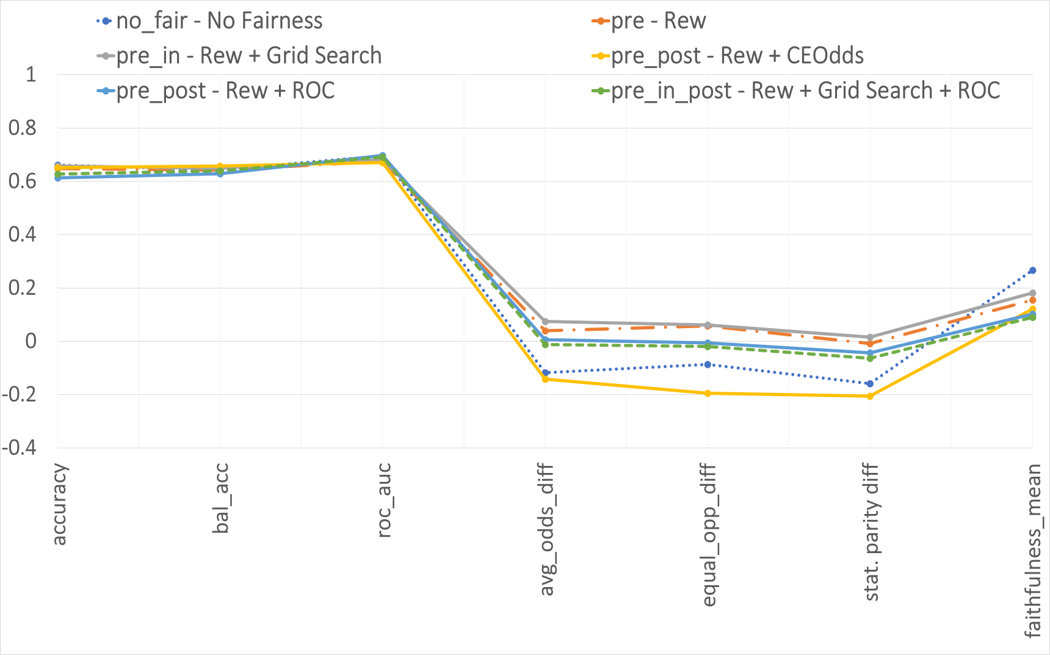}}
\caption{Performance, Fairness, and Interpretability metrics for an XGBoost model learned from the COMPAS dataset after different types of fairness interventions involving a pre-processing bias mitigation stage (protected feature $=$ Race).}
\label{pre_high_DI_Compas-Race}
\end{figure}

\begin{figure}
\centerline{\includegraphics[scale=0.35]{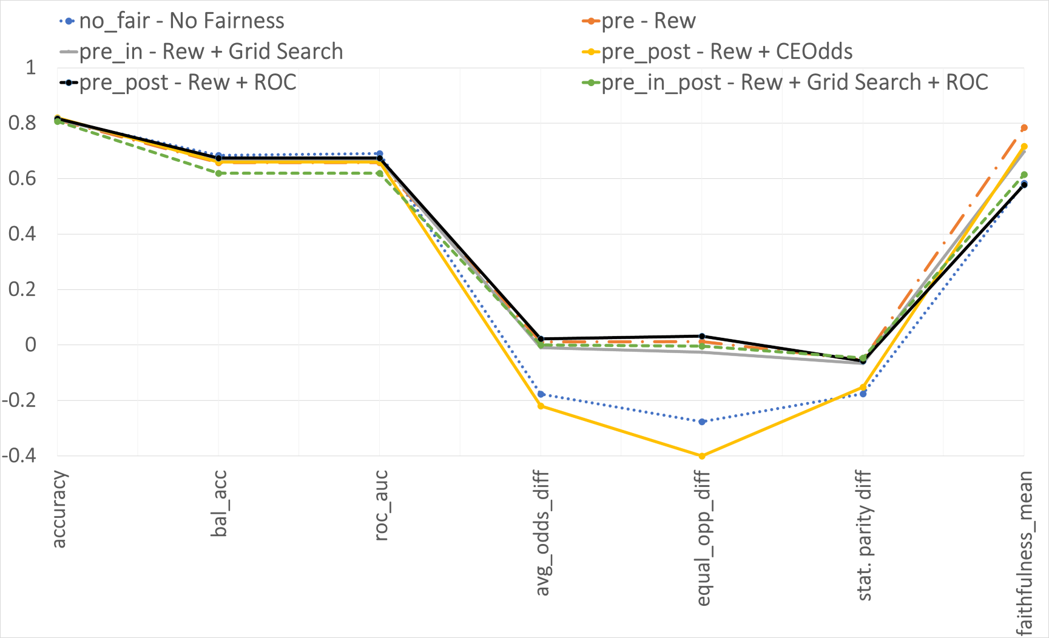}}
\caption{Performance, Fairness, and Interpretability metrics for an MLP model learned from the Adult dataset after different types of fairness interventions involving a pre-processing bias mitigation stage (protected feature $=$ Sex).}
\label{pre_low_DI_Adult-sex}
\end{figure}

Fig. \ref{pre_high_DI_Compas-Race} and Fig. \ref{pre_low_DI_Adult-sex} show the effect of (multiple) bias mitigation interventions on the performance of resultant models, as well as the quality of post-hoc local explanations generated from such models.  Fig. \ref{pre_high_DI_Compas-Race} shows various metrics for different kinds of fairness pipelines involving a pre-processing stage (\texttt{Rew}) for the COMPAS dataset, where the base (XGBoost) model before pre-processing was already quite fair (DI approximately 0.8) and resulted in a fair model after pre-processing intervention. On the other hand, Fig. \ref{pre_low_DI_Adult-sex} shows results for the Adult dataset, where the base (MLP) model was very unfair (DI $<$ 0.3) and even pre-processing mitigation (\texttt{Rew}) could not increase it beyond 0.8. In both cases, performance (accuracy/balanced accuracy/ROC AUC) takes a hit; however, the negative impact is much more in the case of Adult, since multiple interventions are needed to make the resultant model fair. 

Three additional fairness metrics, average odds difference, equal opportunity difference, and statistical parity difference, are also shown. While all pipelines improve fairness in both cases, simple pre-processing is enough in the case of COMPAS (as the model was already quite fair); however, additional mitigation interventions helped make the model fairer in the case of Adult.

Finally, the quality of the post-hoc local explanations generated using LIME for the model after various interventions is shown by measuring the faithfulness of such explanations for a random set of instances. Fairness interventions result in a deterioration in the quality of the local explanations generated. Just introducing a single pre-processing stage had the least effect; additional steps caused further deterioration in the quality of the explanations.

\subsection{In-/Post-Processing-Only Pipelines}

\begin{figure}
\centerline{\includegraphics[scale=0.35]{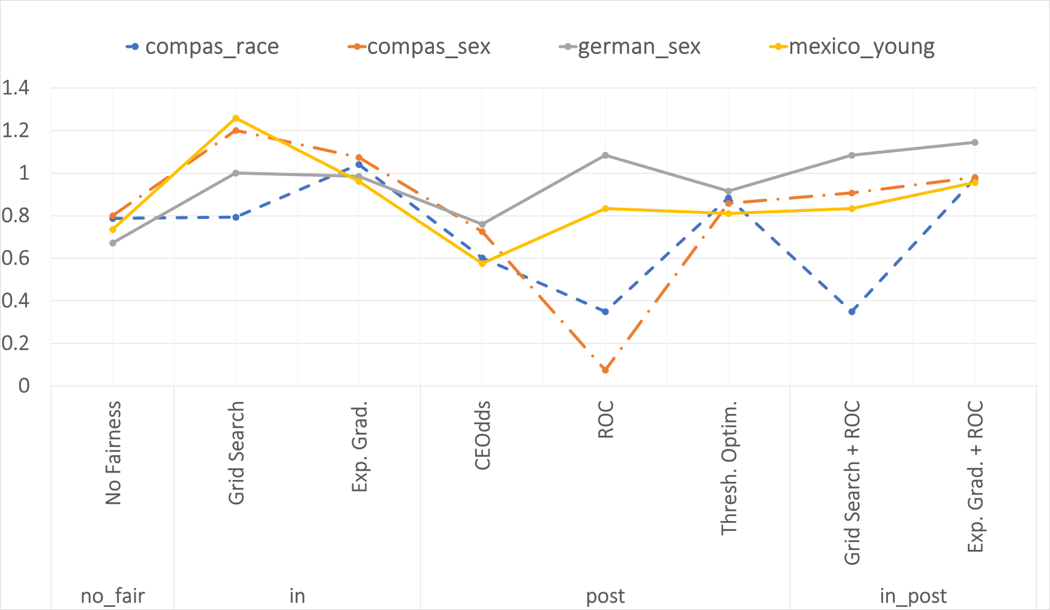}}
\caption{Disparate impact (DI) values for an XGBoost model after different types of fairness interventions without a pre-processing bias mitigation stage where DI value of base model before bias mitigation $>$ 0.6.}
\label{in_post_high_DI}
\end{figure}

\begin{figure}
\centerline{\includegraphics[scale=0.35]{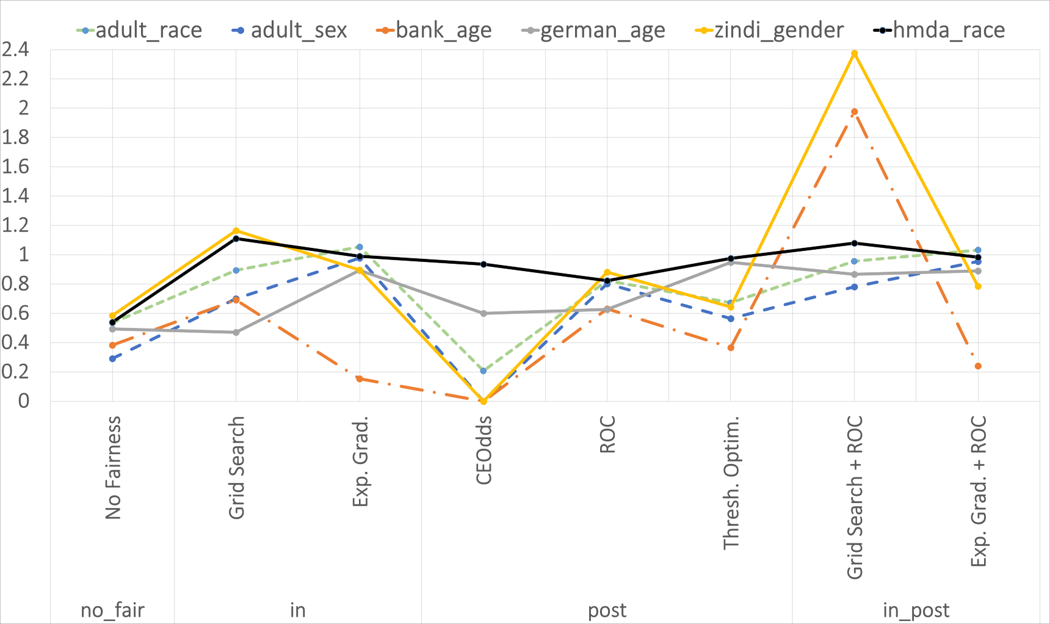}}
\caption{Disparate impact (DI) values for an XGBoost model after different types of fairness interventions without a pre-processing bias mitigation stage where DI value of base model before bias mitigation $<$ 0.6.}
\label{in_post_low_DI}
\end{figure}

\begin{figure}
\centerline{\includegraphics[scale=0.35]{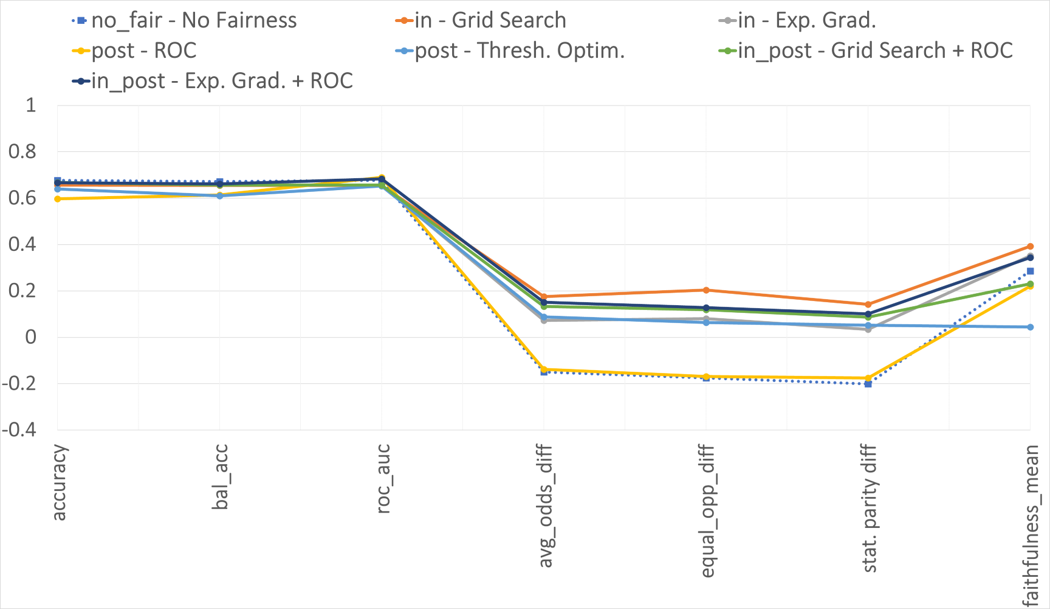}}
\caption{Performance, Fairness, and Interpretability metrics for an MLP model learned from the COMPAS dataset after different types of fairness interventions without a pre-processing bias mitigation stage (protected feature $=$ Sex).}
\label{in_post_high_DI_Compas-Sex}
\end{figure}

\begin{figure}
\centerline{\includegraphics[scale=0.35]{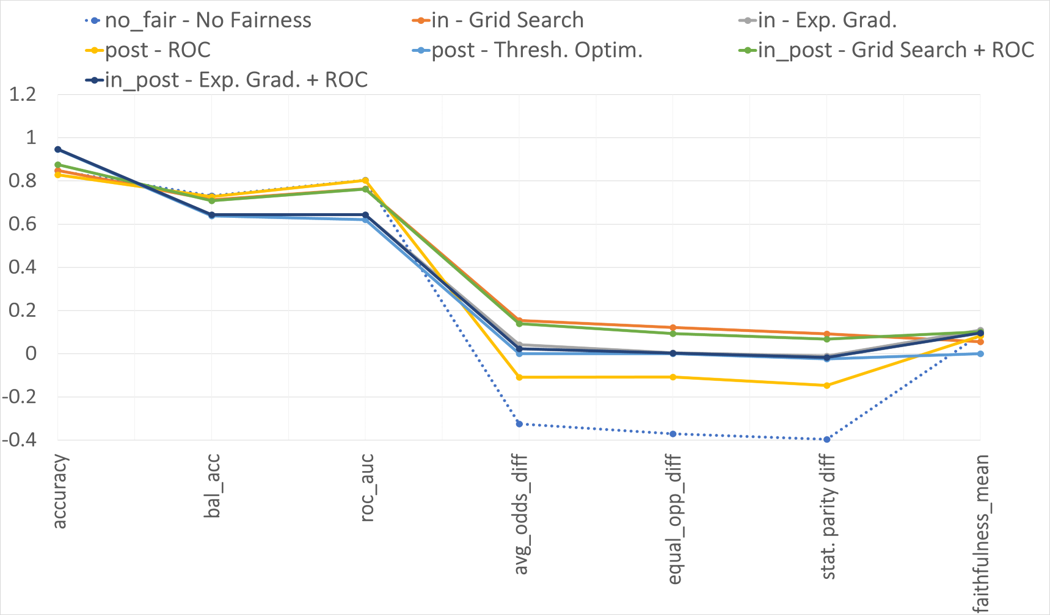}}
\caption{Performance, Fairness, and Interpretability metrics for an XGBoost model learned from the HMDA dataset after different types of fairness interventions without a pre-processing bias mitigation stage (protected feature $=$ Race).}
\label{in_post_low_DI_HMDA-race}
\end{figure}

Fig. \ref{in_post_high_DI}--\ref{in_post_low_DI_HMDA-race} show the results of similar experiments involving pipelines that do not include a pre-processing intervention. Such situations can arise, for example, in situations where certain policies or regulations do not allow training data to be transformed in any way. In such cases, fairness can only be achieved using in-processing and post-processing mitigation techniques. While Fig. \ref{in_post_high_DI} shows results for datasets for which the base model had disparate impact $>$ 0.6, Fig. \ref{in_post_low_DI} shows corresponding results for the datasets which had relatively less fair base models (DI $<$ 0.6). In both cases, in-processing intervention alone worked reasonably well in improving the fairness of the models. However, in cases where the base model DI was relatively high to begin with ($>$ 0.8), intervention with \texttt{GridSearch} tended to over-compensate and resulted in models biased in the opposite direction. Moreover, post-processing interventions alone produced fairly inconsistent results; as in the case of pipelines involving pre-processing, \texttt{CEOdds} and \texttt{ThreshOptim} often performed worse than in-processing. More importantly, the addition of a post-processing stage to a pipeline in conjunction to an in-processing stage showed fairly dataset dependent results. In some cases, the fairness of the model was improved (such as \texttt{GridSearch} + \texttt{ROC} for German, Adult, and HMDA, all cases where the bias model was fairly unfair) while in other cases the fairness was wildly overcompensated (such as Bank and Zindi with \texttt{GridSearch} + \texttt{ROC}, even though the base Zindi model was quite unfair). Thus, although it seems that using a pipeline of in-processing and post-processing interventions can be helpful in learning fair models in situations where the base model is very unfair, it may be more dataset dependent compared to pipelines involving a pre-processing stage.

Effects of fairness interventions using pipelines of in-processing and post-processing interventions on model performance and explanation quality are shown in Fig. \ref{in_post_high_DI_Compas-Sex} for COMPAS, and Fig. \ref{in_post_low_DI_HMDA-race} for the HMDA dataset. As in the case of pipelines involving a pre-processing stage, performance here too is negatively impacted by fairness interventions. In all cases (except, as pointed out above, for the \texttt{ROC}-only pipeline for COMPAS), the pipelines resulted in improved fairness over the base model. The quality of the local explanations generated by LIME too deteriorated, with the greatest effect being felt in the case of pipelines with both in-processing and post-processing interventions.

\begin{figure}
\centerline{\includegraphics[scale=0.25]{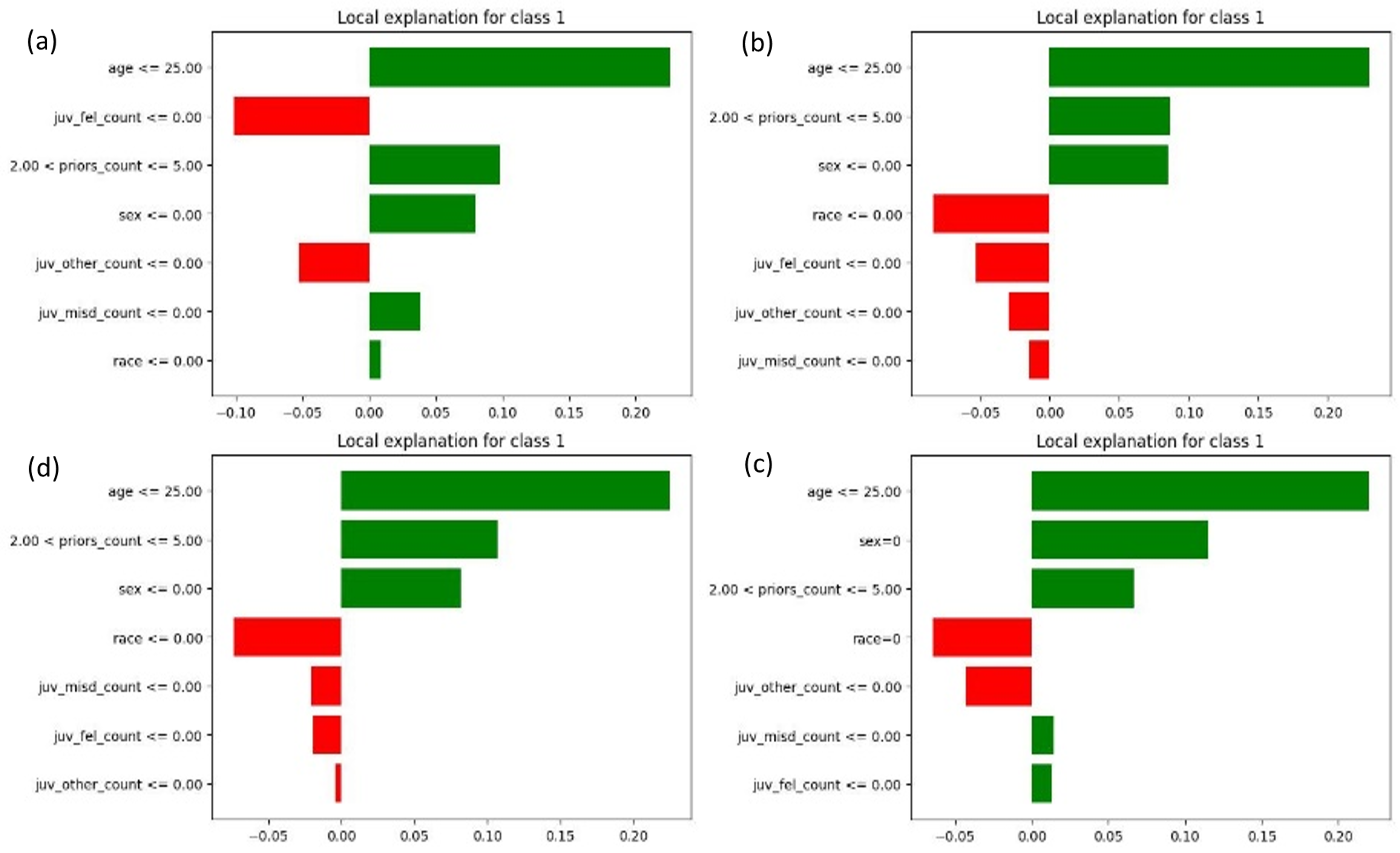}}
\caption{Local post-hoc explanations generated by LIME for predictions made by XGBoost model on the same data instance in the COMPAS dataset for a pipeline with a pre-processing stage (\texttt{Rew}), an in-processing stage (\texttt{GridSearch}), and a post-processing stage (\texttt{ROC}). The explanations shown are generated from model (in clockwise order) (a) before any intervention, (b) after \texttt{Rew}, (c) after \texttt{Rew} + \texttt{GridSearch}, and (d) after \texttt{Rew} + \texttt{GridSearch} + \texttt{ROC}. The protected attribute was race, the person is question was Black (unprivileged group), and the true label/prediction were both Recidivate = 1.}
\label{compas_race_Rew+GridSearch+ROC_explanations}
\end{figure}

By generating explanations from a model after every fairness intervention, we are also able to see what impact the interventions have on the explanations. Fig. \ref{compas_race_Rew+GridSearch+ROC_explanations} shows the local post-hoc explanations generated by LIME at every stage of a pipeline with pre-processing (\texttt{Rew}), in-processing (\texttt{GridSearch}), and post-processing (\texttt{ROC}) interventions for an instance from the COMPAS dataset using the XGBoost model. The particular instance in this case refers to a Black individual, and the protected attribute in this case was race. As can be seen from Fig. \ref{pre_high_DI}, the base model in this case has a fairly high DI value (approximately 0.8) meaning that it is slightly biased against the unprivileged group (Black in this case). Moreover, just the pre-processing intervention alone resulted in an almost fair model, with little impact on fairness from the resulting stages. This is borne out by the LIME explanations. 
Fig. \ref{compas_race_Rew+GridSearch+ROC_explanations} shows how post-hoc explanations change for a particular instance as fairness mitigation steps are applied, one after another. The explanations change to show the effect that the sensitive feature (race) has on predictions.
The base model (top-left) shows that race being Black slightly increases the chances of the model predicting the person to recidivate, as would be expected from a model that is biased against Black individuals. A \texttt{Reweighing} intervention adjusts the model bias by both increasing the importance of the Race variable, as well moving the bias in the opposite direction (i.e., race = Black reduces the chance of the model predicting the person as recidivating) to counter the bias in the base rates of the data. 
The magnitude of this effect increases as more mitigations are applied. 
The polarity of the other features, such as age, sex, and priors remains unchanged during the various fairness interventions, as would be expected in this particular case.

\begin{figure}
\centerline{\includegraphics[scale=0.35]{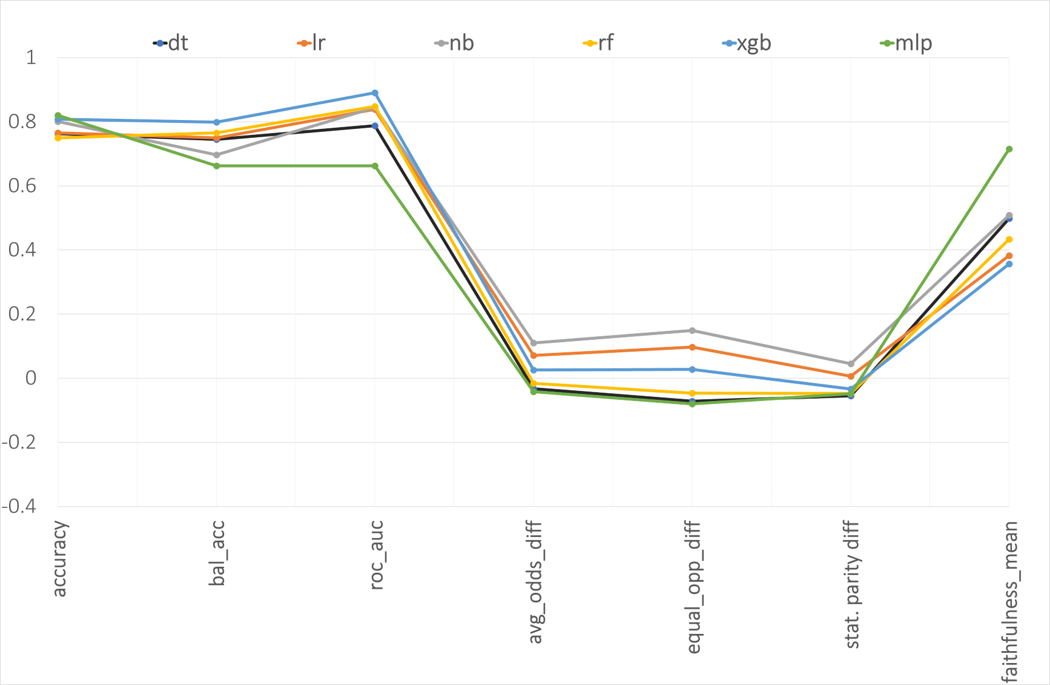}}
\caption{Performance, Fairness, and Interpretability metrics for various models learned from the Adult dataset after fairness intervention using a pipeline  comprising a pre-processing stage (Rew), an in-processing stage (GridSearch), and a post-processing stage (ROC) (protected feature $=$ Race).}
\label{all_models_perf_adult_race}
\end{figure}

\begin{figure}
\centerline{\includegraphics[scale=0.35]{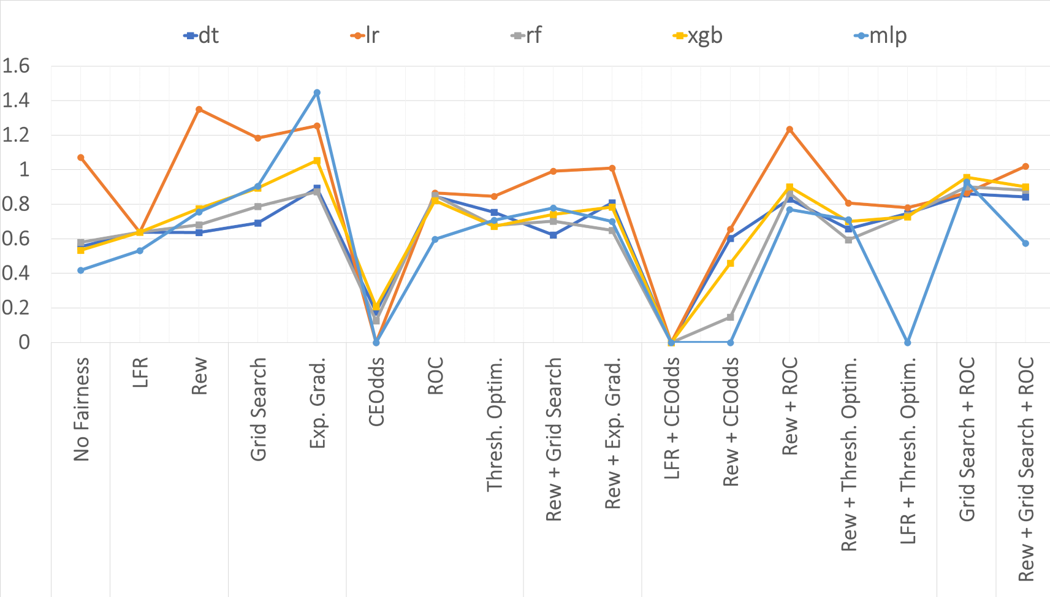}}
\caption{Disparate impact values for various models learned from the Adult dataset after different types
of fairness interventions, with protected feature $=$ Race.}
\label{all_models_adult_race}
\end{figure}

For conciseness and clarity, we chose to present results of experiments primarily for only one model class, XGBoost, as it typically was the best performer on most of the datasets. In a few cases, we have shown results for a Multi Layer Perceptron model instead, to demonstrate the validity of observations on this class of models too. Various model types showed similar behaviors in terms of fairness, performance, and explainability when subjected to fairness interventions using different kinds of pipelines, although the magnitude of the impact depended on the dataset and model type. Figure \ref{all_models_perf_adult_race} shows performance, fairness, and explainability metrics for different types of models learned from the Adult dataset after intervention with a \texttt{Rew} + \texttt{GridSearch} + \texttt{ROC} pipeline, while Figure \ref{all_models_adult_race} shows how the fairness of different models (as measured by disparate impact) changed when subjected to different bias mitigation pipelines.

\section{Conclusion and Future Work}
\label{sec:conclusion}

In this work, we investigate the composition of functions arising in different pillars of trustworthiness. We conduct experiments on several real-world datasets to gain insights on the most effective combinations and pipelines. Thus far, we have focused on bias mitigation algorithms applicable in the three different parts of an ML pipeline and on post-hoc explanations. Furthermore, We report progress on an extensible \emph{composer tool} that allows users to string together functions from multiple pillars of trustworthiness. 

Some novel insights that these experiments have already shown include the following. The quality of explanations (as given by the faithfulness metric) deteriorates as more fairness interventions are applied. Moreover, there is no ideal composition of fairness interventions for any general scenario. The resultant impacts on performance, fairness, and explainability are largely model- and dataset-dependent, with best practices related to the initial unfairness present in a base model applied to the dataset. Importantly, these insights are only illustrative and only scratch the surface of what is possible to learn by experimenting with combinations of several interventions.

In the future, we plan to add functionalities from other dimensions of trustworthiness as well as continue to add other bias mitigation algorithms and explainability methods (from all parts of the pipeline, not just post-processing). We aim to integrate methods for adversarial robustness from the Adversarial Robustness Toolbox \cite{nicolae2018adversarial}, uncertainty calibration and quantification from Uncertainty Quantification 360 \cite{ghosh2021uncertainty}, and privacy and distributional robustness from various toolkits. Ultimately, we hope our composer tool encourages the holistic consideration of multiple pillars of trustworthiness when dealing with machine learning problems in practice.

\bibliographystyle{ACM-Reference-Format}
\bibliography{sample-base}

\appendix

\section{Supplemental Material}
Table \ref{tab1} on the next page summarizes the datasets used in our experiments.

\begin{table*}[h]
\caption{Summary of datasets used in experiments.}
\begin{center}
\begin{tabular}{crrccl}
\toprule
Dataset & Size & Features & Protected Attribute(s) & Privileged Group & Favorable Label \\
\midrule
Adult Income & 48,442 & 7 & race, sex & White, male & 1 (income $>$ \$50K) \\
Bank Marketing & 43,354 & 7 & age & age $< 25$ & 1 (subscriber) \\
COMPAS & 5,278 & 7 & race, sex & Caucasian, female & 0 (no recidivism) \\
German Credit & 1,000 & 9 & age, sex & age $\ge 25$, male & 1 (good credit risk) \\
HMDA & 1,119,631 & 35 & race-ethnicity & non-Hispanic Whites & 1 (mortgage approved) \\
Mexico Poverty & 70,305 & 186 & age & split by mean age & 1 (not impoverished) \\
Zindi & 23,524 & 38 & gender & male & 1 (owns bank account) \\
\bottomrule
\end{tabular}
\label{tab1}
\end{center}
\end{table*}

\end{document}